\documentclass[acmtog,nonacm]{acmart}
\AtBeginDocument{%
  \providecommand\BibTeX{{%
    \normalfont B\kern-0.5em{\scshape i\kern-0.25em b}\kern-0.8em\TeX}}}

\setcopyright{none}
\copyrightyear{2021}
\acmYear{}
\acmDOI{}
\acmISBN{}

\acmConference[DLG '21]{DLG '21: The Sixth International Workshop on Deep Learning on Graphs: Methods and Applications}{August 14--18, 2021}{Singapore}
\acmBooktitle{DLG '21: The Sixth International Workshop on Deep Learning on Graphs: Methods and Applications, August 14--18, 2021, Singapore}

\begin{document}

\title{$\text{Diff}^2\text{Dist}$: Learning Spectrally Distinct Edge Functions, with Applications to Cell Morphology Analysis}

\author{CB Scott}
\authornote{ These authors contributed equally to this research.\\ Corresponding author:{\texttt{scottcb @ uci . edu}}.}
\email{scottcb@uci.edu}
\orcid{0000-0002-5561-2368}
\author{Eric Mjolsness}
\authornotemark[1]
\email{emj@uci.edu}
\orcid{0000-0002-9085-9171}
\affiliation{%
  \institution{University of California, Irvine}
  \streetaddress{6086 Donald Bren Hall}
  \city{Irvine}
  \state{California}
  \country{USA}
  \postcode{92697-3425}
}

\author{Diane Oyen}
\affiliation{%
  \institution{Los Alamos National Lab}
  \streetaddress{P.O. Box 1663 Bikini Atoll Road, SM-30}
  \city{Los Alamos}
  \state{New Mexico}
  \country{USA}
  \postcode{87545}}
\orcid{0000-0002-1353-3688}
\email{doyen@lanl.gov}

\author{Chie Kodera}
\orcid{0000-0002-3078-8930}
\author{David Bouchez}
\orcid{0000-0003-3545-4339}
\author{Magalie Uyttewaal}
\orcid{0000-0003-2881-6637}
\affiliation{
    \institution{INRAE}
    \streetaddress{Route de St Cyr (RD10)}
    \city{Versailles}
    \postcode{78026 Versailles Cedex}
    \country{France}
}

\renewcommand{\shortauthors}{Scott et al.}

\begin{abstract}
  We present a method for learning ``spectrally descriptive'' edge weights for graphs. We generalize a previously known distance measure on graphs (Graph Diffusion Distance), thereby allowing it to be tuned to minimize an arbitrary loss function. Because all steps involved in calculating this modified GDD are differentiable, we demonstrate that it is possible for a small neural network model to learn edge weights which minimize loss. GDD alone does not effectively discriminate between graphs constructed from shoot apical meristem images of wild-type vs. mutant \emph{Arabidopsis thaliana} specimens. However, training edge weights and kernel parameters with contrastive loss produces a learned distance metric with large margins between these graph categories. We demonstrate this by showing improved performance of a simple k-nearest-neighbors classifier on the learned distance matrix. We also demonstrate a further application of this method to biological image analysis: once trained, we use our model to compute the distance between the biological graphs and a set of graphs output by a cell division simulator. This allows us to identify simulation parameter regimes which are similar to each class of graph in our original dataset. 
\end{abstract}



\begin{teaserfigure}
  \includegraphics[width=\textwidth]{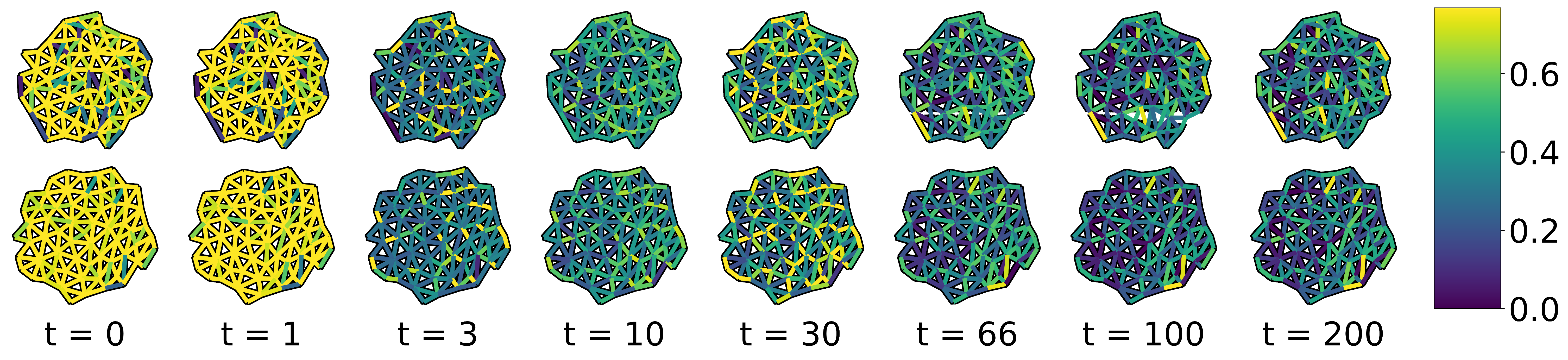}
  \caption{A neural network model learns edge weights which distinguish two classes of graphs. Each row shows the weight values assigned by the network at different times during the training process, from pre-training (far left) to convergence (right). The top row represents a patch of wild-type \emph{Arabidopsis} cells, and the bottom row represents mutants. The pictured edge weights cause these two categories of graph to have distinct spectra.}
  \Description{A neural network model learns edge weights which distinguish two classes of graphs. Each row shows the edge weight values assigned by the network at different times during the training process, from pre-training (far left) to convergence (right). The top row represents a patch of wild-type \emph{Arabidopsis} cells, and the bottom row represents mutants. The pictured edge weights cause these two categories of graph to have distinct spectra.}
  \label{fig:teaser}
\end{teaserfigure}

\maketitle              

\section{Introduction}
Graph Diffusion Distance (GDD) is a measure of similarity between graphs originally introduced by Hammond et. al \cite{hammond2013graph} and greatly expanded by Scott et. al \cite{scott2021graph}. This metric measures the similarity of two graphs by comparing their respective spectra (the eigenvalues of the graph Laplacian). However, it is well-known that there exist pairs of \emph{cospectral} graphs which are not isomorphic but have identical spectra. Furthermore, because even a small change to entries of a matrix may change its eigenvalues, another limitation of GDD is that it is sensitive to small changes in the topology of the graph (as well as small variations in edge weights). Finally, since GDD does not make use of edge or node attributes, it cannot distinguish between two different signals on the same source graph, diminishing its applicability in data science. In this work, we provide several generalizations to GDD which resolve these issues and make it a powerful machine learning tool for datasets of graphs.
\section{Graph Diffusion Distance}
\label{sec:gdd}
We use the definition of Graph Diffusion Distance (GDD) first given by Hammond et. al and later expanded (to cover differently-sized graphs) by Scott et. al. Given two graphs $G_1$ and $G_2$, let $L_1$ and $L_2$ be their respective graph Laplacians. Furthermore, let $L_i = U_i \Lambda_i U_i^T$ be the diagonalizations of each Laplacian, so that $\Lambda_i$ is a diagonal matrix whose $j$th diagonal entry, $\lambda^{(i)}_j$, is the $j$th eigenvalue of $L_i$ . Then the graph diffusion distance between these graphs is given by 
\begin{align}
    \label{eqn:diff_dist_defn}
    D(G_1,G_2) 
    &= \sup_t {\left| \left| e^{t L_1} - e ^ {t L_2} \right| \right|}_F 
    = \sup_t {\left| \left|  e^{t \Lambda_1} - e^{t \Lambda_2} \right| \right|}_F \nonumber \\
    &= \sup_t \sqrt{\sum_{j=1}^n {\left( e^{t \lambda^{(1)}_j} - e^{t \lambda^{(2)}_j} \right)}^2 } .
\end{align}
This simplification relies on several properties of the Frobenius norm and the exponential map (rotation invariance and continuity, respectively) which we shall not detail here. It is clear that this distance measure requires the two graphs to be the same size, since otherwise this matrix difference is not defined. 

The generalization to different-sized graphs given by Scott et. al can also be modified in the way we discuss in Section \ref{sec:learned_gdd}, but we do not consider this version of GDD in this paper. Given two graphs $G_1$, $G_2$ of differing sizes $n_1 < n_2$, we can define graph diffusion distance similarly to Equation \ref{eqn:diff_dist_defn}:
\begin{align}
    \label{eqn:diff_dist_defn_diff_size}
    D(G_1,G_2) 
    &= \sup_{t > 0} \inf_{\alpha > 0} \inf_{P | \mathcal{C}(P)} {\left| \left| P e^{\frac{t}{\sqrt{\alpha}} L_1} - e ^ {t \sqrt{\alpha} L_2} P \right| \right|}_F \nonumber \\
    &= \sup_{t > 0} \inf_{\alpha > 0} \inf_{\tilde{P} | \mathcal{C}(\tilde{P})} {\left| \left| \tilde{P} e^{\frac{t}{\sqrt{\alpha}} \Lambda_1} - e ^ {t \sqrt{\alpha} \Lambda_2} \tilde{P} \right| \right|}_F. 
\end{align}
In Equation \ref{eqn:diff_dist_defn_diff_size}, $\alpha$ is a time-dilation factor which dilates the passage of time in one graph with respect to the other. P is a rectangular matrix which is optimized according to some set of constraints $\mathcal{C}$. In the cited paper by Scott et. al, $\mathcal{C}$ is taken to be orthogonality: $P^T P = I$. $\tilde{P} = U_2^T P U_1$ is a change of basis from graph-space to eigenspace, allowing us to  again represent the equation for varying-size GDD as a comparison between lists of eigenvalues.

\subsection{GDD is a differentiable function of $t$ and edge weights}
\label{subsec:differentiable}
Once all of the eigenvalues $\lambda_i$ and eigenvectors $v_i$ (of a matrix $L$) are computed, we may backpropagate through the eigendecomposition as described in \cite{nelson1976simplified} and \cite{andrew1993derivatives}. If our edge weights (and therefore the values in the Laplacian matrix $L$) are parametrized by some value $\theta$, and our loss function $\mathcal{L}$ is dependent on the eigenvalues of $L$, then we can collect the gradient $\frac{\partial \mathcal{L}}{\partial \theta}$ as:
\begin{equation}
    \label{eqn:derivative_eigen}
    \frac{\partial \mathcal{L}}{\partial \theta} = \sum_k \left( \frac{\partial \mathcal{L}}{\partial \lambda_k} \frac{\partial \lambda_k}{\partial \theta} \right) = \sum_k \left( \frac{\partial \mathcal{L}}{\partial \lambda_k} v_k^T \frac{\partial L}{\partial \theta} v_k \right).
\end{equation}
In practice, if the entries of $L$ are computed as a function of $\theta$ using an automatic differentiation package (such as PyTorch \cite{neurips2019pytorch}) the gradient matrix $\frac{\partial L}{\partial \theta}$ is already known before eigendecomposition. We note here that for any fixed value of $t$, all of the operations needed to compute GDD are either simple linear algebra or continuous or both. Therefore, for any loss function $\mathcal{L}$ which takes the GDD between two graphs as input, we may optimize $\mathcal{L}$ by backpropagation through the calculation of GDD using Equation \ref{eqn:derivative_eigen}. We note here that although all numerical experiments in this paper use the same-sized version of GDD, this backpropagation will work for the varying-sized version as well, allowing gradients to be used to adjust any of the inputs to the GDD equation. 

\section{Learning Parameters for Diffusion Kernels}
\label{sec:learned_gdd}
In this section, we describe our method for learning edge weights for Laplacian diffusion kernels, beginning with our generalization of GDD to make it trainable, and then introducing a method for learning edge weights.
\subsection{Diff2Dist: Differentiable Graph Diffusion Distance}
\label{subsec:weighted_gdd}
We make two main changes to GDD to make it capable of being tuned to specific graph data. First, we replace the real-valued optimization over $t$ with a maximum over an explicit list of $t$ values ${t_1, t_2, \ldots t_p}$. This removes the need for an optimization step inside the GDD calculation. Second, we re-weight the Frobenius norm in the GDD calculation with a vector of weights $\beta_j$ which is the same length as the list of eigenvalues (these weights are normalized to sum to 1). The resulting GDD calculation is then:
\begin{equation}   
\label{eqn:diff_dist_tunable_v1}
    D(G_1,G_2) 
    = \max_{t \in t_1, t_2, \ldots t_p} \sqrt{\sum_{j=1}^n {\beta_j \left(e^{t \lambda^{(1)}_j} - e^{t \lambda^{(2)}_j} \right)}^2 } .
\end{equation}
We call this version of GDD \emph{Differentiable Graph Diffusion Distance}, or Diff2Dist. Because this distance calculation is comprised entirely of differentiable components and linear algebra, it may be explicitly included in the computation graph (e.g. in PyTorch) of a machine learning model, without needing to invoke some external optimizer to find the supremum over all $t$. $t_n$ and $\beta_j$ may be tuned by gradient descent or some other optimization algorithm to minimize a loss function which takes $d$ as input. Tuning the $t_n$ values results in a list of values of $t$ for which GDD is most informative for a given dataset, while tuning $\beta_j$ reweights GDD to pay most attention to the eigenvalues which are most discriminative. In the experiments in Section \ref{sec:numeric} we demonstrate the efficacy of tuning these parameters using contrastive loss. 
\subsection{Learning Edge Weighting Functions}
\label{subsec:learn_edge_weights}
Here, we note that if graph edge weights are determined by some differentiable function $f$ parametrized by parameters $\theta$, we may still apply all of the machinery of Sections  \ref{subsec:differentiable} and \ref{subsec:weighted_gdd}. A common edge weighting function for graphs embedded in Euclidean space is the \emph{Gaussian Distance Kernel}, $w_{ij} = \exp \left(
\frac{-1}{2\sigma^2} d_{ij}\right)$, where $d_{ij}$ is the distance between nodes $i$ and $j$ in the embedding. $\sigma$ is the `radius' of the distance kernel and can be chosen \emph{a priori} or tuned in the same way as $\beta$ and $t$, using a numerical optimization procedure. However, the Gaussian distance kernel, while mathematically well-motivated for embedded graphs, is a somewhat arbitrary choice of edge weight, especially in cases where the edges of our graphs have more complicated edge labels. In cases like the data discussed in this paper, our edge labels are vector-valued, and it is therefore advantageous to replace this hand-picked edge weight with weights chosen by a general function approximator, e.g. an artificial neural network \cite{fukushima1982neocognitron}. As before, the parameters of this ANN could be tuned using gradients backpropagated through the GDD calculation and eigendecomposition. 
\section{Numeric Experiments}
\subsection{Data Description}
%
%
\subsubsection{\emph{Arabidopsis} Cell Graph Dataset}
\label{sec:arabid}
The species \emph{Arabidopsis thaliana} is of high interest in plant morphology studies, since 1) its genome was fully sequenced in 1996, relatively early \cite{kaul2000analysis}, and 2) its structure makes it relatively easy to capture images of the area of active cell division: the \emph{shoot apical meristem (SAM)}. Recent work \cite{schaefer2017preprophase} has found that mutant \emph{Arabidopsis} specimens with a loss of function of genes \textit{trm6},\textit{trm7}, and \textit{trm8} demonstrate more variance in the placement of new cell walls during cell division. This is thought to be the result of the \textit{trm}-mutants having abnormal or absent pre-prophase bands (PPBs) during cell divison. The PBB is a cellular substructure which fine-tunes the placement of cell walls during division.

\begin{figure}
    \centering
    \null \hfill \includegraphics[width=.47\linewidth]{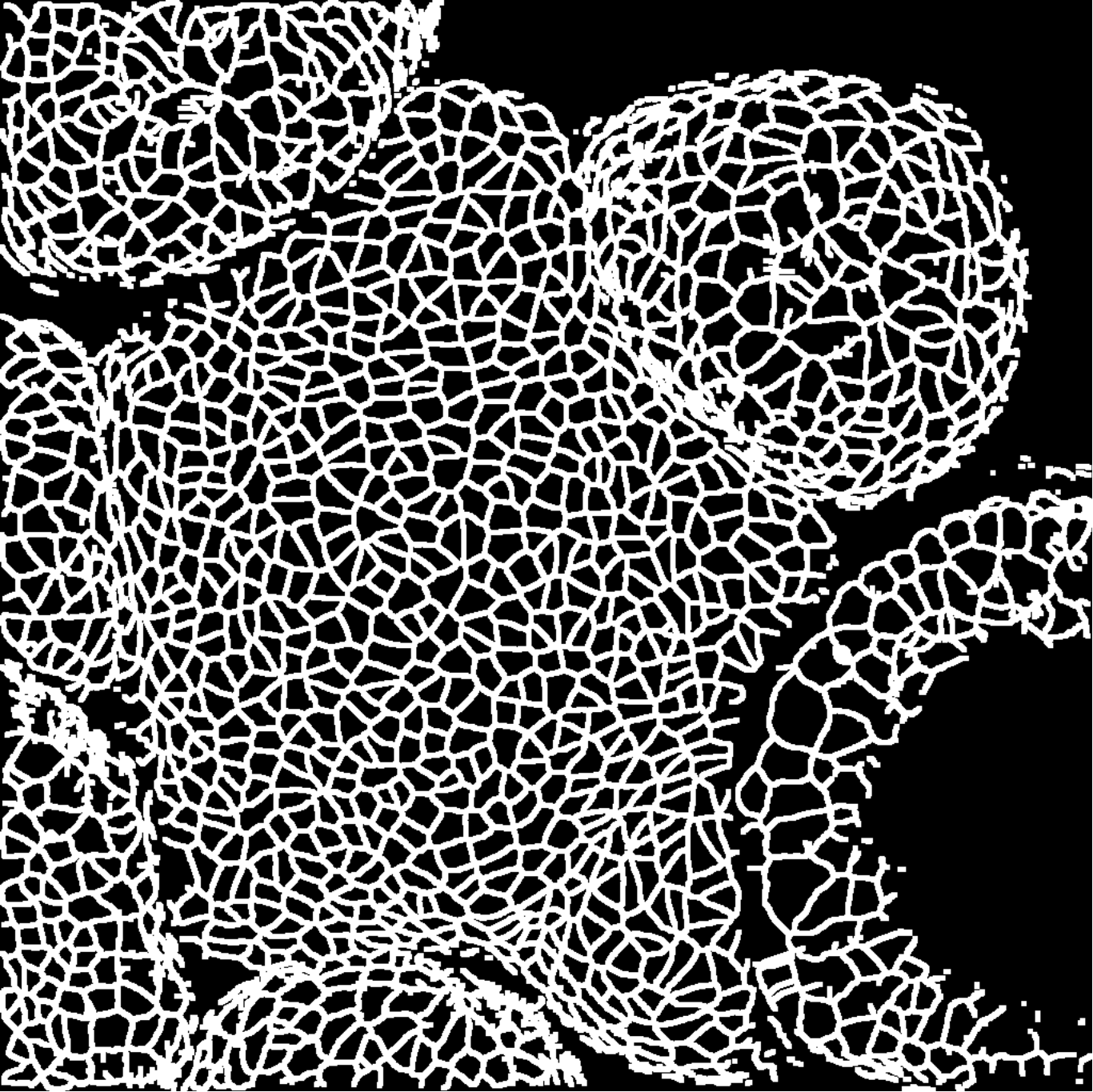} \hfill 
    \includegraphics[width=.47\linewidth]{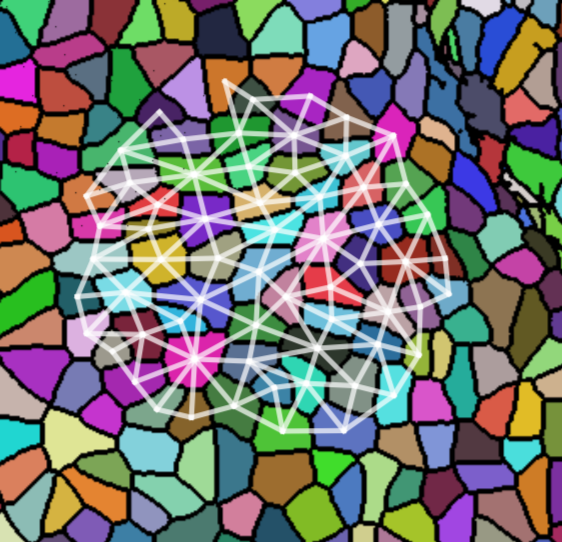} \hfill \null
    \caption{Left: an image of the SAM of a mutant \emph{Arabidopsis} specimen. The original 3D confocal microscope image is here represented as a 2D skeleton. Right: a zoomed-in view of the same specimen, with separate cells false-colored and an example extracted cell neighborhood graph overlaid.}
    \label{fig:cell_lines}
\end{figure}

We take 20 confocal microscope images (13 from wild-type plants and 7 from \textit{trm678} mutants) of shoot apical meristems, and process them to extract graphs representing local neighborhoods of cells. Each graph consists of a cell and its 63 closest neighbors (64 cells total). Cell neighborhood selection was limited to the central region of each SAM image, since the primordia surrounding the SAM are known to have different morphological properties. For each cell neighborhood, we produce a graph by connecting two cells if and only if their shared boundary is 30 pixels or longer  (pixel size : 1 px = 0.1818 µm). For each edge, we save the length of this shared boundary, as well as the angle of the edge from horizontal and the edge length. We extracted 2318 cell neighborhoods in this way, resulting in a dataset of 13. See Figure \ref{fig:cell_lines} for an example morphological graph. 
\label{sec:numeric}
We test each of the GDD generalizations proposed, on the task of classifiying wild-type vs. mutant morphological graphs. We split our dataset 85\%/15\% train/validation; all metrics we report are calculated on the validation set. We compare the following four methods: \begin{enumerate}
    \item Original GDD of unweighted graphs, with no tuning of $t$ or other parameters;
    \item Diff2Dist of graphs with Gaussian kernel edge weights (fixed $\sigma$), with $t$ and $\beta_i$ tuned;
    \item Diff2Dist of graphs with Gaussian kernel edge weights, with $t$, $\sigma$, and $\beta_i$ all tuned;
    \item Diff2Dist of graphs with general edge weights parametrized by a small neural network. Input to this neural network was a vector of all three edge attributes. 
\end{enumerate}
For methods 2 and 3, the input to the distance kernel was the distance between nodes in the original image. All parameters were tuned using ADAMOpt \cite{kingma2014adam} (with default PyTorch hyperparameters and batch size 256) to minimize the \emph{contrastive loss} function \cite{hadsell2006dimensionality}:
\begin{align}
    \label{eqn:contrast_loss} \mathcal{L}(G_i, G_j) &= \frac{1}{2} \left( y_{ij}{\max(0, d(G_i, G_j)  \rho_\text{lower} )}^2 \right. \nonumber \\ 
    &\left. + (1-y_{ij}){\max(0, \rho_\text{upper} - d(G_i, G_j) )}^2 \right). 
\end{align}
This loss function encourages $G_i$ and $G_j$ to be closer than $\rho_\text{lower}$ if they have the same label, and further apart than $\rho_\text{upper}$ if they differ ( $y_{ij}$ is a binary indicator of label agreement). These margins were set ($\rho_\text{lower}=0.001$, $\rho_\text{upper} = 0.33$) by trial-and-error on the training set. Training took 600 epochs. For the neural network approach, edge weights were chosen as the final output of a neural network with seven layers of sizes $\{3,128,32,32,32,32,1\}$ with SiLU activations on the first six layers and no activation function on the last layer.   
Results of these experiments can be found in Table \ref{tab:valid_results_morpho}. We also present distance matrices for each approach, as well as Isomap \cite{tenenbaum2000global} embeddings of each (we used the scikit-learn implementation of Isomap with 15 neighbors and default hyperparameters). The distance matrices developed using the ANN approach clearly show better separation between the two categories. The distance matrices and resulting embeddings for methods 1 and 2 do not result in a clear separation between the two categories, while methods 3 and 4 do. Furthermore, method 4 (Diff2Dist) demonstrates a dramatic improvement over method 3, showing that this version of GDD can be tuned so that the resulting graph distance minimizes an arbitrary loss function (e.g. separates classes of graphs).
%
%
\begin{table}
    \centering
    \begin{tabular}{|c | c | c|}
    \hline
        Method & Accuracy \% \\
        \hline
         GDD only & 89.4  \\
         $t$-tuning and $\beta$-weights & 90.6 \\ 
         $t$ and $\sigma$-tuning,$\beta$-weights & 94.7 \\ 
         ANN Parametrization & \textbf{99.1} \\
         \hline
    \end{tabular}
    \caption{Validation set accuracy for a simple K-nearest neighbors classifier for all four methods. The validation set was the same for each of these tests. The value reported is the highest value over all $K\in\{3, 50\}$. }
    \label{tab:valid_results_morpho}
\end{table}
\begin{figure}[ht]
    \centering
    \includegraphics[width=.6\linewidth]{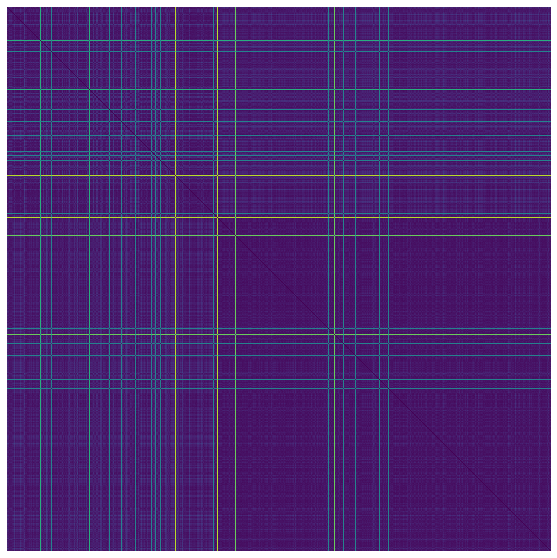} \\
    \includegraphics[width=.6\linewidth]{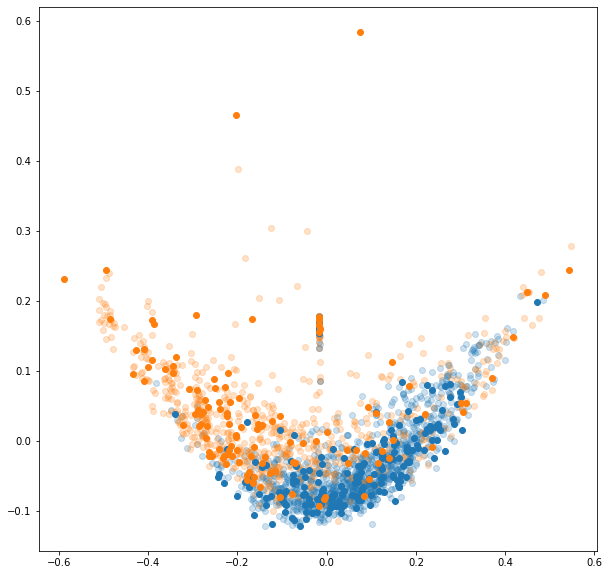}
    \caption{Top: distance matrix for morphological graphs, generated with method 1 (GDD on unweighted graphs). Bottom: Isomap embedding of this distance matrix, which ensures that points with small distance are placed near each other. Blue points represent wild type, and orange points represent mutants. Transparent points represent training set graphs; solid points represent those from the validation set. We see that naive GDD leads to an embedding where the two categories of graph overlap, indicating that GDD by iteself is not capturing the distinction between these two classes of graph.}
    \label{fig:expt1_fig}
\end{figure}

\begin{figure}[ht]
    \centering
    \includegraphics[width=.6\linewidth]{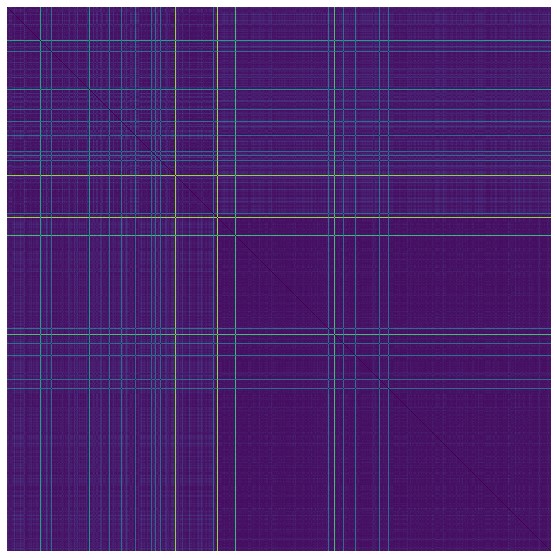} \\
    \includegraphics[width=.6\linewidth]{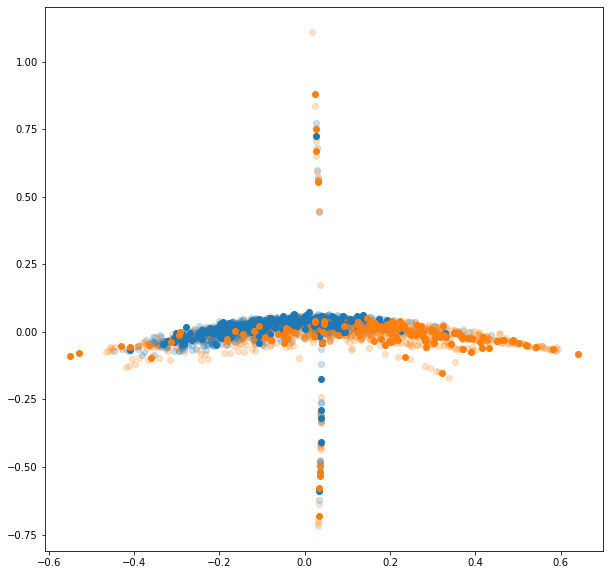}
    \caption{As in Figure \ref{fig:expt1_fig}, but for method 2 (parametrized GDD on unweighted graphs). This distance matrix and embedding have the same flaws as those in Figure \ref{fig:expt1_fig}.}
    \label{fig:expt2_fig}
\end{figure}

\begin{figure}[ht]
    \centering
    \includegraphics[width=.6\linewidth]{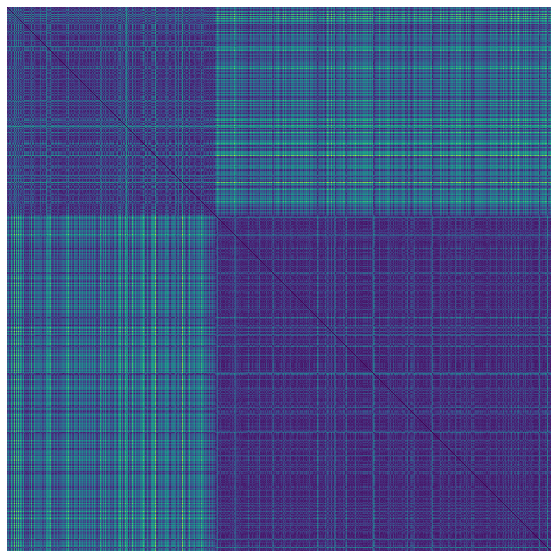} \\
    \includegraphics[width=.6\linewidth]{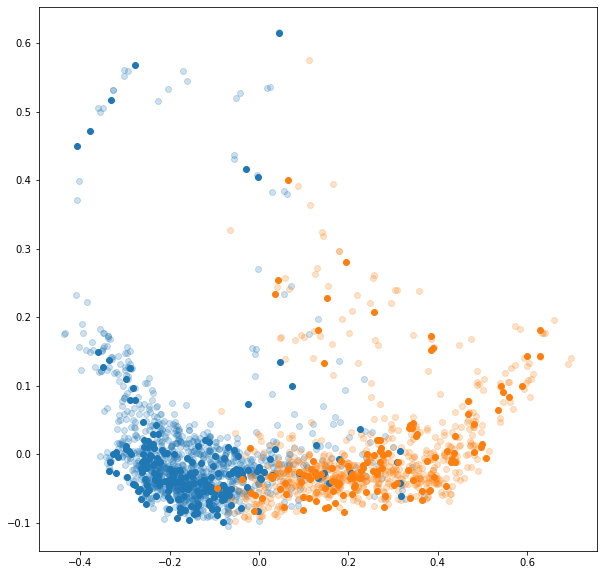}
    \caption{As in Figure \ref{fig:expt1_fig}, but for method 3 (Gaussian kernel with tuned $\sigma$). We see that using a Gaussian distance kernel for edge weights, with radius tuned by gradient descent, results in a graph distance metric which better separates the two categories.}
    \label{fig:expt3_fig}
\end{figure}

\begin{figure}[ht]
    \centering
    \includegraphics[width=.6\linewidth]{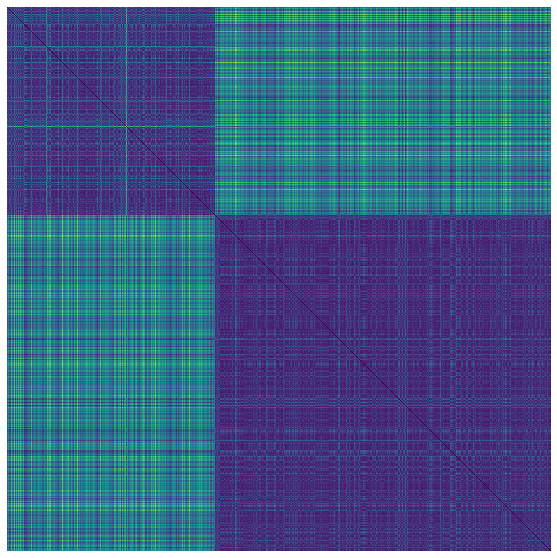} \\
    \includegraphics[width=.6\linewidth]{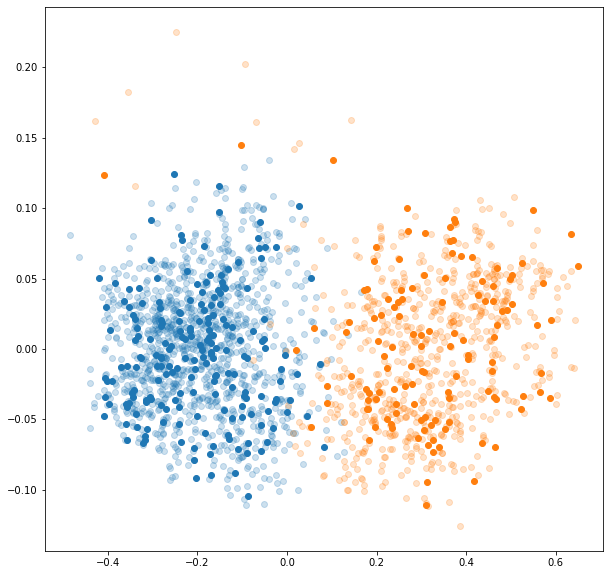}
    \caption{As in Figure \ref{fig:expt1_fig}, but for the ANN-determined edge weights. Replacing the arbitrary Gaussian distance kernel with weights chosen by a machine-learning model makes this approach fully general and produces a distance metric which fully separates the two graph categories.}
    \label{fig:expt4_fig}
\end{figure}
\section{Analysis of Simulation Hyperparameters}
In this section, we use the distance metric trained in the previous section to characterize the differences in cell growth between the two types of cells, by comparing the biological graphs to graphs generated by a simulation. 

We generated a population of artificial morphological graphs using the simulation code Tissue \cite{Hamant1650}. In our model, cells divide when they reach a volume of 40 arbitrary units, with a small random chance $r_\text{div}$ of dividing at any timestep if they are larger than 20 units. Cell division planes are usually placed along the shortest path between two non-adjacent cell walls, but are randomly oriented with probability $r_\text{angle}$. These parameters were implemented in a new division rule we contributed to the Tissue simulator (our modified version of Tissue's source code is available in the supplementary material of this manuscript). We also varied two parameters from the original version of Tissue, representing the spring constant of each cell wall and the exclusion radius around each vertex (where new vertices are not allowed to be placed during division). A summary of the values swept over for each parameter is in Table \ref{tab:sim_params}. There were $4 \times 3 \times 4 \times 6 = 288$ combinations of parameters, resulting in that many simulations (all simulations were started from the meristem.init file included with Tissue). Each simulation resulted in a mesh file representing the final positions of cell vertices and cell walls after 10000 timesteps. Each mesh was converted into an image and processed into a set of graphs (one centered on each cell) exactly as described in Section \ref{sec:arabid}.

\begin{figure}[h]
    \centering
    \includegraphics[width=.47\linewidth]{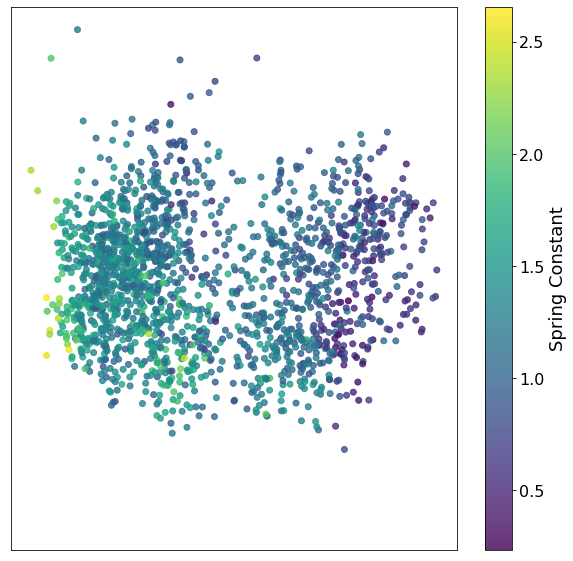} \hfill
    \includegraphics[width=.47\linewidth]{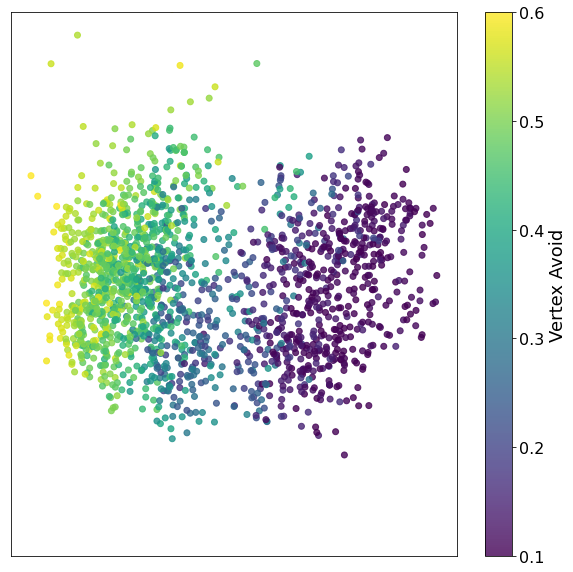} \\
    \includegraphics[width=.47\linewidth]{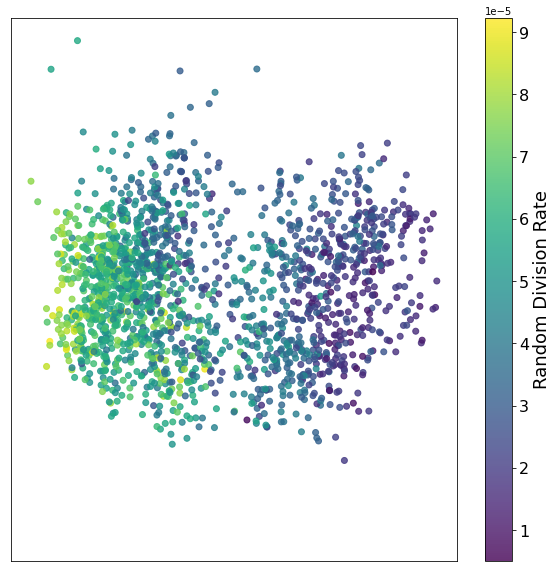} \hfill
    \includegraphics[width=.47\linewidth]{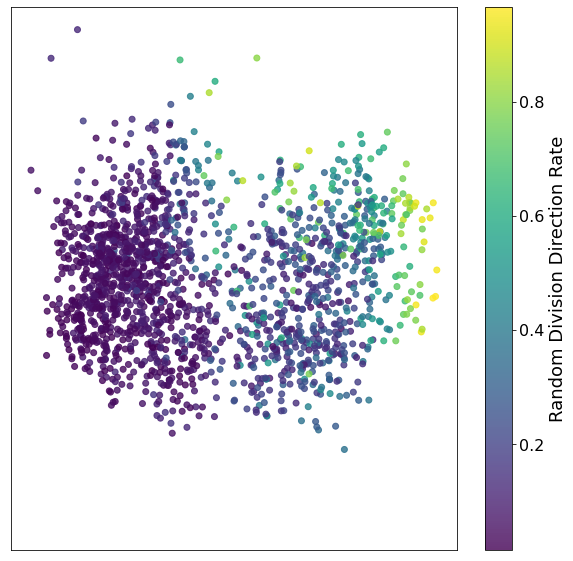}
    \\
    
    \caption{Visualizing simulation parameters using Diff2Dist. Each plot shows one point for each morphological graph in our \emph{Arabidopsis} dataset. Points are placed in 2D using Isomap, exactly as in Figure \ref{fig:expt4_fig}. Points are colored according to the parameter values of the nearest simulation-derived graphs, where 'nearest' means in the sense of our trained distance metric. }
    \label{fig:param_viz}
\end{figure}

We used the Diff2Dist model (trained on the biological graphs only) to compute the distance between each biological graph and all of the graphs which originated from a simulation. Each biological graph can then be assigned a numerical label for each parameter, by taking the mean of the parameters for the 100 closest simulations. This gives us an estimate of which parameter values a given biological graph is most similar to (under our learned distance estimate, which is shown to separate \textit{trm678} and wild-type graphs). The results of this experiment can be seen in Figure \ref{fig:param_viz}. Comparing biologically derived graphs to synthetically-generated ones in this way allows us to see that wild-type graphs are characterized by high vertex avoidance and low rate of random cell division direction, while \textit{trm678} graphs are more likely to have cell divisions in random directions.    

\begin{table}
    \centering
    \begin{tabular}{|c|c|}
    \hline
     Parameter Name  & Values used \\
     \hline
     Spring Constant    & 0.1, 0.3, 1.0, 3.0 \\
     Vertex Exclusion Size & 0.1, 0.3, 0.6 \\
     Random Div Freq $r_\text{div}$ & 0.0, 0.00001, 0.00003, 0.0001 \\
     Random Div Direction $r_\text{angle}$ & 0.0, 0.01, 0.03, 0.1, 0.5, 1.0 \\
    \hline
    \end{tabular}
    \caption{Summary of input parameters and values used during comparison of simulations to biologically-derived graphs. Spring constant controls the stiffness of cell walls during the simulation. Vertex exclusion size is the proportional size of an envelop around each vertex where new vertices may not be placed during division.  $r_\text{div}$ is the random chance of a cell dividing at a given timestep even when it has not reached 40 units of volume.  $r_\text{angle}$ is the probability (given that division has occurred) that the placement of the new cell wall will be random instead of optimal.}
    \label{tab:sim_params}
\end{table}

\section{Conclusion and Future Work}
This paper presents a method to compute distance metrics between edge-labelled graphs, in such a way as to respect class labels. This approach is flexible and can be implemented entirely in PyTorch, making it possible to learn a distance metric between graphs that were previously not able to be discriminated by Graph Diffusion Distance. In the future hope to apply this method to more heterogenous graph datasets by including the varying-size version of GDD. We also note here that our neural network approach, as described, is not a Graph Neural Network in the sense described by prior works like \cite{kipf2016semi,bacciu2020gentle}, as there is no message-passing step. We expect message-passing layers to directly improve these results and hope to include them in a future version of differentiable GDD.

\section{Acknowledgements}
Research presented in this article was supported by the Laboratory Directed Research and Development program of Los Alamos National Laboratory under project number 20200041ER. This work was also supported by the UC Irvine Donald Bren School of Information and Computer Sciences Endeavor grant, and by grants from the Human Frontiers Science Program [grant HFSP - RGP0023/2018] and National Institute of Health [grant R01 HD073179]. The authors would also like to thank (in no particular order) Jaques Dumais, Olivier Hamant, And Henrik J\"onsson for their advice and guidance.

\clearpage
\bibliographystyle{splncs04}
\bibliography{egbib}

\end{document}